**AutoScore-Ordinal: An Interpretable Machine Learning Framework for Generating Scoring Models for Ordinal Outcomes**

Seyed Ehsan Saffari[1,2#], Yilin Ning[1#], Xie Feng[1,2], Bibhas Chakraborty[1,2,3,4], Victor Volovici[5,6], Roger Vaughan[1,2], Marcus Eng Hock Ong[2,7], Nan Liu[1,2,8,9]*

[1] Centre for Quantitative Medicine, Duke-NUS Medical School, Singapore

[2] Programme in Health Services and Systems Research, Duke-NUS Medical School, Singapore

[3] Department of Biostatistics and Bioinformatics, Duke University, Durham, NC, USA

[4] Department of Statistics and Data Science, National University of Singapore, Singapore

[5] Department of Neurosurgery, Erasmus MC Stroke Center, Erasmus MC University Medical Center Rotterdam, Rotterdam, The Netherlands

[6] Department of Public Health, Erasmus University, Rotterdam, The Netherlands

[7] Department of Emergency Medicine, Singapore General Hospital, Singapore

[8] SingHealth AI Health Program, Singapore Health Services, Singapore

[9] Institute of Data Science, National University of Singapore, Singapore

[#] These authors contributed equally.

* Corresponding author: Nan Liu, Centre for Quantitative Medicine, Duke-NUS Medical School, 8 College Road, Singapore 169857, Singapore. Phone: +65 6601 6503. Email: liu.nan@duke-nus.edu.sg

**ABSTRACT**

**Background**

Risk prediction models are useful tools in clinical decision-making which help with risk stratification and resource allocations and may lead to a better health care for patients. AutoScore is a machine learning–based automatic clinical score generator for binary outcomes. This study aims to expand the AutoScore framework to provide a tool for interpretable risk prediction for ordinal outcomes.

**Methods**

The AutoScore-Ordinal framework is generated using the same 6 modules of the original AutoScore algorithm including variable ranking, variable transformation, score derivation (from proportional odds models), model selection, score fine-tuning, and model evaluation. To illustrate the AutoScore-Ordinal performance, the method was conducted on electronic health records data from the emergency department at Singapore General Hospital over 2008 to 2017. The model was trained on 70% of the data, validated on 10% and tested on the remaining 20%.

**Results**

This study included 445,989 inpatient cases, where the distribution of the ordinal outcome was 80.7% alive without 30-day readmission, 12.5% alive with 30-day readmission, and 6.8% died inpatient or by day 30 post discharge. Two point-based risk prediction models were developed using two sets of 8 predictor variables identified by the flexible variable selection procedure. The two models indicated reasonably good performance measured by mean area under the receiver operating characteristic curve (0.785 and 0.793) and generalized c-index (0.737 and 0.760), which were comparable to alternative models.

**Conclusion**

AutoScore-Ordinal provides an automated and easy-to-use framework for development and validation of risk prediction models for ordinal outcomes, which can systematically identify potential predictors from high-dimensional data.

# INTRODUCTION

Risk prediction models are mathematical equations which help clinicians estimate the probability of a healthcare outcome, given patient data. Such models include integer-point scores which can be used to predict that a disease is present (diagnostic models) or a specific outcome will occur (prognostic models), depending on the clinical question. A combination of multiple predictors (different weights for different predictors) is included into a multivariable model to calculate a risk score (1–3). Some risk prediction models have been used in routine clinical settings, including the Framingham Risk Score (4), Ottawa Ankle Rules (5), Nottingham Prognostic Index (6), Gail model (7), Euro-SCORE (8), the modified Early Warning Score (MEWS) (9,10) and Simplified Acute Physiology Score (11).

The use of health information technology, particularly electronic health records (EHR), has increased in the past decade, which provides opportunities for big data research. EHR data includes detailed patient information and clinical outcome variables which can be a unique data source for risk model development (12,13). Availability of a large number of variables in EHR data could be a mathematical challenge when using traditional regression analysis to build up a risk model. Machine learning (ML), as an alternative approach, applies mathematical algorithms to handle such big data resulting in novel risk prediction models. Traditional variable selection approaches (such as backward elimination, forward selection, stepwise selection using pre-specified stopping rules) may result in different subsets of variables in the context of EHR data, and clinical knowledge might not be always available in some clinical domains. Powerful feature selection techniques are available for supervised learning, which is a very critical aspect in risk model development when working with EHR data (13,14).

AutoScore (15) is an easy-to-use, machine learning–based automatic clinical score generator, which develops interpretable clinical scoring models. In an empirical experiment using EHR data, AutoScore generated scoring models that achieved comparable predictive performance as several conventional methods for risk model development but by using fewer variables (15). The advantage of the AutoScore framework is the combination of efficient variable selection using ML techniques and the accessibility and interpretability of simple regression models. It can be easily used in different clinical settings and its applicability has been shown with a large number of variables (EHR data, for example) (15). Some recent studies have used this framework to develop a risk prediction model in various clinical domains (16,17).

Most risk prediction models in the literature were developed using multivariable logistic regression models or ML techniques to predict a binary outcome. Many clinical ordinal outcome variables exist and they are often dichotomized (favorable and unfavorable) for simplicity. Nevertheless, it should not be ignored that such re-categorization results in loss of clinically and statistically relevant information, which may also involve difficulties in borderline patients (cases that can easily be categorized into either of the two levels of the outcome). One should note that analyzing ordinal variables has more statistical power in comparison to the corresponding re-categorized binary variables. This has been illustrated in both simulations and empirical studies in clinical trials (18–22). Literature also recommends the use of the ordinal scale outcomes rather than dichotomization, as smaller treatment effect sizes are detectable via ordinal analysis (19,23–25).

In the literature, ordinal outcome variables are discussed in several clinical domains, where the objective was either an association exploration or predictions. A large international study (including 26 hospitals from six countries) conducted ordinal logistic regression to study a composite ordinal outcome variable (defined as 1=alive, no long length of stay [LOS], no readmission; 2=alive, long LOS, no readmission; 3=alive, no long LOS, readmission; 4 = alive, long LOS, readmission; 5= death), and the correlation among different levels of the composite ordinal outcome at hospital level was reported (26). Machine learning methods using multiple biomarkers were performed to develop an ovarian cancer–specific predictive framework in a retrospective cohort study of 435 patients on a secondary ordinal outcome of residual tumor size (defined as: no residual tumor, <1 cm residual tumor, ≥1 cm residual tumor), and the predictive accuracy and AUC were discussed (27). However, none of the above studies have generated an interpretable risk scoring system which could be easily used in the clinic for real-time decision making.

There is a lack of literature in model development using ordinal analysis which can be easily applied to clinical studies dealing with complex data (EHR, for example). The primary objective of this study was to expand the original AutoScore framework to provide a tool for easy development and validation of risk prediction models for ordinal outcomes. For illustration purpose, a risk prediction model was developed and validated using EHR data from the emergency department, where the ordinal outcome included three categories (alive without readmission to the hospital within 30 days post discharge, alive with readmission within 30 days post discharge and dead inpatient or within 30 days post discharge).

**METHODS**

**AutoScore-Ordinal Framework**

As in the original AutoScore framework (15), in Module 1 (see Figure 1) the data is first split into a training set to train prediction models, a validation set to select hyper-parameters (e.g., number of variables, cut-off values for categorizing continuous variables), and a test set to evaluate the final model(s) selected. The three datasets typically contain 70%, 10% and 20% of the full dataset, respectively. Variables are ranked based on their importance to a random forest (RF) (28) for multiclass classification (i.e., ignoring the ordering of categories), trained on the training set with a default number of 100 trees.

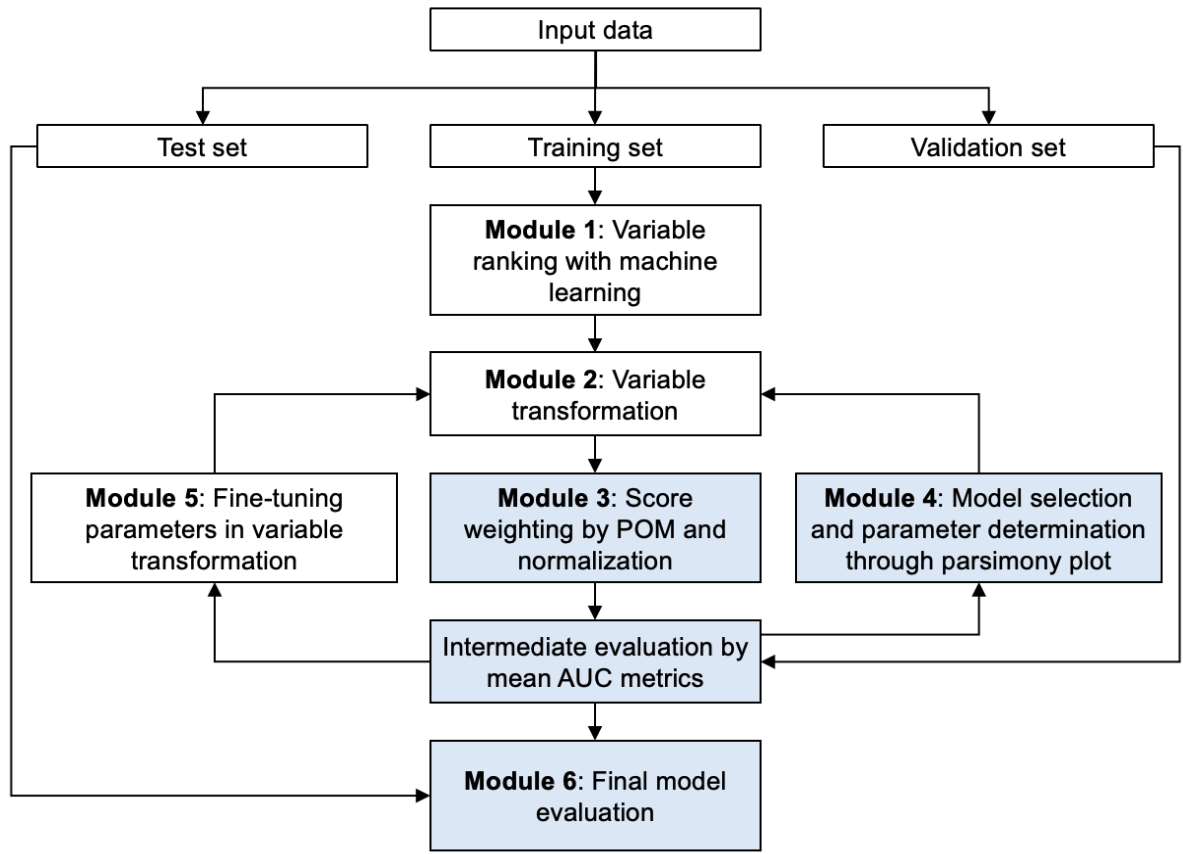

**Figure 1.** Visual illustration of the AutoScore-Ordinal workflow. Blue color highlight modules modified from the original AutoScore framework (15).

To simplify the interpretation and account for possible non-linear relationship between the predictor variables and the outcome, all continuous variables are categorized in Module 2 (see Figure 1). As described in the original AutoScore framework (15), a common practice in clinical applications is to use the 5-th, 20-th, 80-th and 95-th percentiles of a continuous variable (based on the training set) as cut-off values, but some cut-offs may be removed to avoid sparsity issues when the distribution of a variable is highly skewed.

In Module 3 (see Figure 1), weights associated with variables are developed using the proportional odds model (POM) (29,30), which is one of the most widely used regression models in studies of ordinal outcomes. Let scalar $Y$ denote the ordinal outcome with $J$ categories (denoted by integers $1, ..., J$) and column vector $x$ denote the variables (with continuous variables readily categorized in Module 2). The POM assumes a linear model for the logit of the cumulative probabilities associated with the $j$-th ordinal category, i.e., $p_j = P(Y \leq j), j = 1, \ldots, J - 1$:

$$\log\left(\frac{p_j}{1 - p_j}\right) = \theta_j - x^T \beta.$$

The scalar terms $\theta_j$ are category-specific intercept terms, where $\theta_1 < \theta_2 < \cdots < \theta_{J-1}$ to ensure $p_j < p_k$ for any $j < k$. $\beta$ is the vector of regression coefficients corresponding to the predictors. The negative sign before $\beta$ follows from the notation used by McCullagh (29,30), such that a positive value of $\beta$ indicates a positive association between $x$ and $Y$, i.e., an increase in $x$ leads to an increased probability of observing a higher category in $Y$. Hence an increase in $x^T \beta$ is always associated with increased probabilities of observing higher outcome categories, allowing us to construct prediction scores based on $x^T \beta$.

To improve interpretability, the POM is refitted after redefining reference categories in each variable such that all elements in $\beta$ are positive, and $\beta$ is normalised with respect to the minimum value of $\beta$. With all continuous variables readily categorised in Module 2, these normalised coefficients can be interpreted as scores associated with a category of a variable, referred to as partial scores. The partial scores (which are 0 for reference categories and 1 or larger otherwise) are rounded to positive integers to simplify the calculation of final prediction scores, which is the summation of all partial scores corresponding to the values of variables for an individual. To facilitate interpretation, all partial scores are often rescaled (and then rounded) such that the maximum total score attenable is a meaningful value (e.g., 100).

To evaluate the performance of the final model, the prediction of outcome $Y$ with $J$ categories is divided into $J - 1$ binary classifications of $Y \leq j$ vs $Y > j$, and the mean area under the receiver operating characteristic curve (AUC) across these binary classifications (referred to as mAUC hereafter) is used to evaluate the overall performance for predicting $Y$, which is equivalent to the average dichotomized c-index for evaluating ordinal predictions (31,32). In Module 4, a scoring model is grown by adding one variable at each time (based on the variable ranking from Module 1) until all candidate variables are included, and the improvement in mAUC (evaluated on the validation set) with increasing number of variables is inspected using the parsimony plot. The final list of variables is often selected when the benefit of adding a variable is small. Next, the cut-off values for continuous variables selected in Module 4 may be fine-tuned for favourable interpretation in Module 5. The final model is evaluated on the test set in Module 6 using the mAUC and Harrell's generalised c-index (31,33,34), which is based on the proportion of concordant pairs (i.e., when predictions and observed outcomes generate the same ranking for the pair of observations, including tied ranks) among all possible pairs of observations. For both mAUC and generalised c-index, a value of 0.5 indicates a random performance and a value of 1 indicates a perfect predictive performance. The mAUC and generalised c-index from the test set are reported with the bias-corrected 95% bootstrap confidence interval (CI) (35).

**Data example**

We used AutoScore-Ordinal to predict readmission and death (composite outcome) after inpatient discharge, using data collected from patients who visited the emergency department (ED) of Singapore General Hospital in years 2008 to 2017 and were subsequently admitted to the hospital (36). The full cohort included data on 449,593 ED presentation cases. Information on patient demographics, ED administration, inpatient admission, clinical tests and vital signs in ED, medical history and comorbidities was extracted from the hospital electronic health record system (16). We excluded patients aged below 18, resulting in a final sample of 445,989 inpatient cases.

We constructed a composite ordinal outcome with three categories: alive without readmission to the hospital within 30 days post discharge, alive with readmission within 30 days post discharge, died inpatient or within 30 days post discharge. We randomly split the dataset (stratified by outcome categories) into a training set of 70% (n=312,193) cases to train models, a validation set of 10% (n=44,599) cases to perform necessary model fine-tuning for AutoScore-Ordinal, and a test set of 20% (n=89,197) cases to evaluate the performance of the final prediction models. For each case, we extracted the length of stay (LOS) of the previous inpatient admission (missing values were treated as 0 days). Missing values for vital signs or clinical tests were imputed using the median value in the validation set.

We compared the prediction model built using AutoScore-Ordinal with the RF (with 100 trees) and POM with LASSO or stepwise variable selection techniques. For each model, we computed the 95% CI for mAUC and generalized c-index from bootstrap samples of the test set (the number of bootstrap samples was selected as 100 for the demo purposes and can be modified in the AutoScore algorithm). Generalized c-index was computed based on the total score for AutoScore-generated models, the linear predictor excluding intercept terms for POM and the predicted outcome categories for RF.

**Implementation**

All analyses were implemented in R version 4.0.5 (37). Our proposed AutoScore-Ordinal is implemented as an R package, available from https://github.com/nliulab/AutoScore-Ordinal. POM was implemented using the *clm* function from package *ordinal* (38). The *stepAIC* function from package *MASS* (39) was used to perform stepwise variable selection for POM, and the *ordinalNet* function from package *ordinalNet* (40) was used to implement the LASSO approach. The RF was implemented using the *randomForest* function from package *randomForest* (41). The bias-corrected bootstrap CI was implemented using the *bca* function from package *coxed* (42). The generalized c-index was implemented using the *rcorrcens* function from package *Hmisc* (43).

# RESULTS

Among the 445,989 cases, 359,961 (80.7%) were in the first outcome category (i.e., alive without 30-day readmission), 55,552 (12.5%) were in the second category (i.e., alive with 30-day readmission), and 30,476 (6.8%) were in the third category (i.e., died inpatient or by day 30 post discharge). The characteristics of the full cohort are summarized in Table 1. Cases in the 3 outcome categories showed statistical difference in all variables, therefore it is non-trivial to develop a sparse prediction model based on POM.

**Table 1.** Characteristics of cases in the full cohort. Outcome categories 1, 2, and 3 refer to cases that were alive without readmission to the hospital within 30 days post discharge, alive with readmission within 30 days post discharge and dead inpatient or within 30 days post discharge, respectively.

|  | Overall (n=445,989) | Outcome category 1 (alive, no readmission; n= 359,961) | Outcome category 2 (alive, with readmission; n= 55,552) | Outcome category 3 (death; n= 30,476) |
|---|---|---|---|---|
| **Patient demographics** | | | | |
| Age (years; mean (SD)) | 61.66 (18.24) | 60.16 (18.55) | 66.38 (15.86) | 70.84 (13.83) |
| Male (%) | 222644 (49.9) | 177267 (49.2) | 28753 (51.8) | 16624 (54.5) |
| Race (%) | | | | |
|   Chinese | 27471 ( 6.2) | 24615 ( 6.8) | 1958 ( 3.5) | 898 ( 2.9) |
|   Indian | 316474 (71.0) | 250930 (69.7) | 41022 (73.8) | 24522 (80.5) |
|   Malay | 47508 (10.7) | 39606 (11.0) | 5973 (10.8) | 1929 ( 6.3) |
|   Others | 54536 (12.2) | 44810 (12.4) | 6599 (11.9) | 3127 (10.3) |
| **Comorbidity (%)** | | | | |
| Myocardial infarction | 26594 ( 6.0) | 15653 ( 4.3) | 6242 (11.2) | 4699 (15.4) |
| Congestive heart failure | 49575 (11.1) | 32360 ( 9.0) | 11809 (21.3) | 5406 (17.7) |
| Peripheral vascular disease | 25878 ( 5.8) | 16701 ( 4.6) | 6258 (11.3) | 2919 ( 9.6) |
| Stroke | 57730 (12.9) | 41674 (11.6) | 10463 (18.8) | 5593 (18.4) |
| Dementia | 12385 ( 2.8) | 8129 ( 2.3) | 2625 ( 4.7) | 1631 ( 5.4) |
| Pulmonary | 42770 ( 9.6) | 30385 ( 8.4) | 8868 (16.0) | 3517 (11.5) |
| Rheumatic | 6180 ( 1.4) | 4645 ( 1.3) | 1147 ( 2.1) | 388 ( 1.3) |
| Peptic ulcer disease | 17193 ( 3.9) | 11834 ( 3.3) | 3478 ( 6.3) | 1881 ( 6.2) |
| Mild liver disease | 20483 ( 4.6) | 14318 ( 4.0) | 4216 ( 7.6) | 1949 ( 6.4) |
| Severe liver disease | 7119 ( 1.6) | 3863 ( 1.1) | 1906 ( 3.4) | 1350 ( 4.4) |
| Diabetes (without complications) | 55699 (12.5) | 42529 (11.8) | 8756 (15.8) | 4414 (14.5) |
| Diabetes with complications | 104682 (23.5) | 76553 (21.3) | 19987 (36.0) | 8142 (26.7) |
| Paralysis | 24903 ( 5.6) | 17683 ( 4.9) | 4692 ( 8.4) | 2528 ( 8.3) |
| Renal | 91213 (20.5) | 62033 (17.2) | 20290 (36.5) | 8890 (29.2) |
| Cancer (non-metastatic) | 39571 ( 8.9) | 27627 ( 7.7) | 6778 (12.2) | 5166 (17.0) |
| Metastatic cancer | 35225 ( 7.9) | 18469 ( 5.1) | 5683 (10.2) | 11073 (36.3) |
| **ED Admission** | | | | |
| ED LOS (hours; mean (SD)) | 2.86 (1.70) | 2.84 (1.72) | 2.58 (1.62) | 2.12 (1.42) |
| ED Triage code (%) | | | | |
|   P1 | 83221 (18.7) | 59513 (16.5) | 11696 (21.1) | 12012 (39.4) |
|   P2 | 250382 (56.1) | 199708 (55.5) | 33906 (61.0) | 16768 (55.0) |
|   P3 and P4 | 112386 (25.2) | 100740 (28.0) | 9950 (17.9) | 1696 ( 5.6) |
| ED Boarding Time (hours; mean (SD)) | 4.78 (3.81) | 4.80 (3.79) | 4.79 (3.94) | 4.48 (3.89) |
| Consultation Waiting Time (hours; mean (SD)) | 0.77 (0.80) | 0.80 (0.82) | 0.71 (0.72) | 0.53 (0.60) |
| **Inpatient admission** | | | | |
| Day of Week (%) | | | | |
|   Friday | 62453 (14.0) | 50314 (14.0) | 7801 (14.0) | 4338 (14.2) |
|   Monday | 74192 (16.6) | 60142 (16.7) | 9091 (16.4) | 4959 (16.3) |
|   Weekend | 115418 (25.9) | 92387 (25.7) | 14604 (26.3) | 8427 (27.7) |
|   Midweek | 193926 (43.5) | 157118 (43.6) | 24056 (43.3) | 12752 (41.8) |
| Admission Type (%) | | | | |
|   A1 | 16814 ( 3.8) | 14795 ( 4.1) | 1195 ( 2.2) | 824 ( 2.7) |
|   B1 | 37345 ( 8.4) | 32938 ( 9.2) | 2658 ( 4.8) | 1749 ( 5.7) |
|   B2 | 212261 (47.6) | 174752 (48.5) | 23238 (41.8) | 14271 (46.8) |
|   C | 179569 (40.3) | 137476 (38.2) | 28461 (51.2) | 13632 (44.7) |
| Previous LOS (days; mean (SD)) | 3.57 (8.55) | 3.04 (7.90) | 5.34 (10.00) | 6.50 (11.55) |
| **Healthcare utilisation in the previous year** | | | | |
| No. inpatient visits (mean (SD)) | 0.93 (2.21) | 0.62 (1.42) | 2.66 (4.46) | 1.44 (2.17) |
| No. surgery (mean (SD)) | 0.20 (0.74) | 0.15 (0.63) | 0.42 (1.10) | 0.37 (0.99) |
| No. ICU stays (mean (SD)) | 0.02 (0.25) | 0.02 (0.22) | 0.05 (0.35) | 0.05 (0.36) |
| No. HD stays (mean (SD)) | 0.09 (0.47) | 0.07 (0.40) | 0.17 (0.68) | 0.17 (0.69) |
| **Vital sign and clinical tests** | | | | |
| Ventilation (%) | 89 ( 0.0) | 47 ( 0.0) | 7 ( 0.0) | 35 ( 0.1) |
| Resuscitation (%) | 9083 ( 2.0) | 6211 ( 1.7) | 1045 ( 1.9) | 1827 ( 6.0) |
| Pulse, beat/minute (mean (SD))[*] | 82.85 (17.23) | 81.99 (16.73) | 83.35 (17.00) | 92.05 (20.51) |
| Respiration, breath/minute (mean (SD))[*] | 17.84 (1.77) | 17.76 (1.60) | 17.98 (1.81) | 18.59 (2.98) |
| $SpO_2$, % (mean (SD))[*] | 97.98 (3.25) | 98.05 (3.05) | 97.93 (3.12) | 97.34 (5.22) |
| DBP, mmHg (mean (SD))[*] | 71.33 (13.55) | 71.67 (13.36) | 71.14 (13.70) | 67.62 (14.93) |
| SBP, mmHg (mean (SD))[*] | 133.68 (25.53) | 134.05 (25.19) | 135.98 (26.22) | 125.13 (26.63) |
| Bicarbonate, mmol/L (mean (SD))[*] | 22.78 (3.78) | 22.91 (3.55) | 22.55 (3.97) | 21.67 (5.39) |
| Creatinine, μmol/L (mean (SD))[*] | 154.70 (208.52) | 142.72 (196.69) | 213.21 (259.28) | 185.61 (215.82) |
| Potassium, mmol/L (mean (SD))[*] | 4.16 (0.72) | 4.14 (0.69) | 4.23 (0.76) | 4.36 (0.91) |
| Sodium, mmol/L (mean (SD))[*] | 135.02 (5.17) | 135.30 (4.84) | 134.56 (5.39) | 132.76 (7.30) |

[*]Excluding 9365 missing entries for pulse, 10772 missing entries for respiration, 10704 missing entries for $SpO_2$, 5348 missing entries for SBP and DBP, 56857 missing entries for bicarbonate, 56742 missing entries for creatinine, 58747 missing entries for potassium, and 56678 missing entries for sodium.

Outcome categories were compared using Kruskal-Wallis and Chi-square test for continuous and categorical variables, respectively. All tests had p-value<0.001.

DBP: diastolic blood pressure; ED: emergency department; HD: high dependency ward; ICU: intensive care unit; LOS: length of stay; SBP: systolic blood pressure; SD: standard deviation; $SpO_2$: blood oxygen saturation.

**Variable selection**

      The parsimony plot (see Figure 2) suggests a reasonable model of 8 variables: ED LOS, creatinine, ED boarding time, number of visits in the previous year, age, systolic blood pressure (SBP), bicarbonate and pulse, which reached a mAUC that is only 7.9% lower than that the scoring model using all 41 variables. We refer to this model as Model 1. When using the parsimony plot to select variables, researchers are not restricted to consecutively select

variables in the descending order of importance. For example, we built an alternative model (i.e., Model 2) with 8 variables, where we excluded the 3rd variable (i.e., ED boarding time) from Model 1 that had little impact on mAUC, and added the 14th variable (i.e., metastatic cancer) that incremented the mAUC by approximately 4% when it entered the prediction model.

**Figure 2.** Parsimony plot by the mean area under the curve (mAUC) on the validation set.

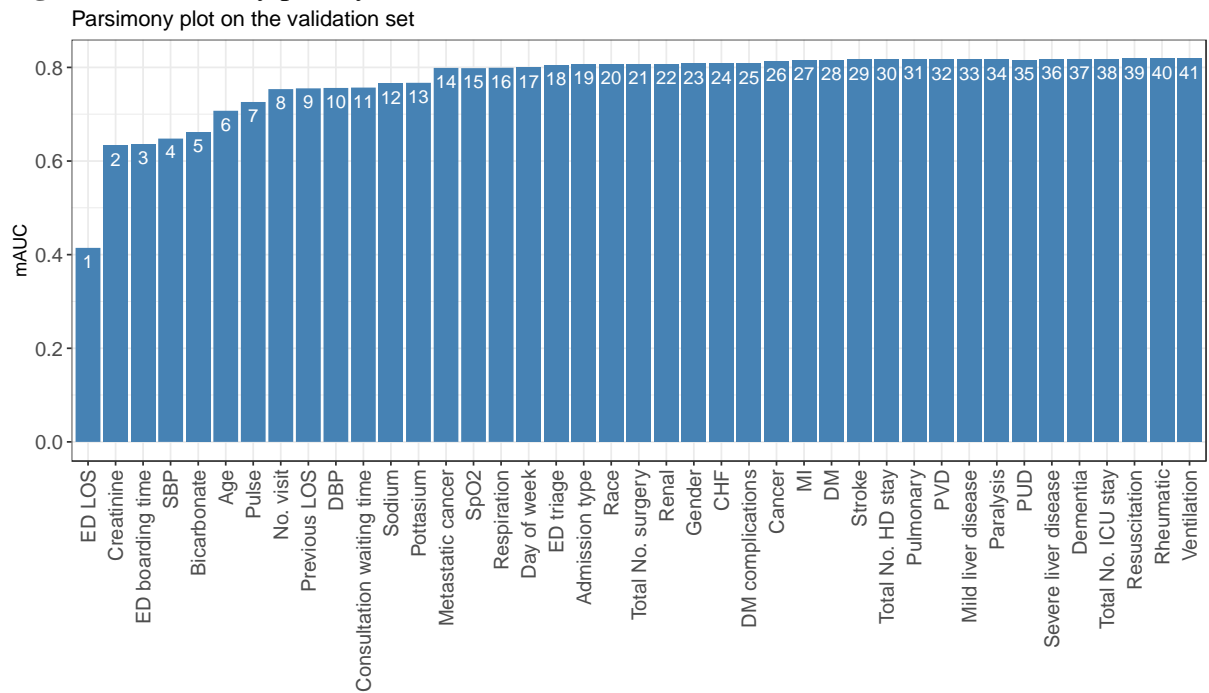

**Fine-tuning**

All variables selected in the two models were continuous, and we fine-tuned their cut-off values in the categorization step to improve interpretability. The scoring tables after fine-tuning were shown in Table 2 for both models, and the performance of the resulting prediction models (evaluated on the test set) were reported in Table 3. Model 1 had an mAUC of 0.757 (95% CI: 0.749-0.762), and by excluding ED boarding time and adding metastatic cancer, the mAUC of Model 2 improved to 0.793 (95% CI: 0.787-0.798).

**Table 2.** Scoring table for AutoScore-generated models.

| Variable | Interval | Partial score for Model 1 | Partial score for Model 2 |
|---|---|---|---|
| ED LOS | <40min | 11 | 7 |
| | [40min, 80min) | 8 | 6 |
| | [80min, 4h) | 4 | 2 |
| | [4h, 6h) | 1 | 1 |
| | >=6h | 0 | 0 |
| Creatinine, μmol/L | <45 | 6 | 4 |
| | [45, 60) | 0 | 0 |
| | [60, 135) | 0 | 1 |
| | [135, 600) | 6 | 7 |
| | >=600 | 4 | 7 |
| ED boarding time | <80min | 0 | -- |
| | [80min, 2.5h) | 2 | -- |
| | >=2.5h | 1 | -- |
| Bicarbonate, mmol/L | <17 | 8 | 7 |
| | [17, 20) | 3 | 3 |
| | [20, 28) | 0 | 0 |
| | >=28 | 5 | 4 |
| Systolic blood pressure, mmHg | <100 | 12 | 9 |
| | [100, 110) | 7 | 5 |
| | [110, 150) | 3 | 2 |
| | [150, 180) | 1 | 0 |
| | >=180 | 0 | 0 |
| Age, years | <25 | 0 | 0 |
| | [25, 45) | 6 | 5 |
| | [45, 75) | 18 | 13 |
| | [75, 85) | 22 | 17 |
| | >=85 | 25 | 21 |
| Pulse, beat/minute | <70 | 0 | 0 |
| | [70, 95) | 3 | 2 |
| | [95, 115) | 8 | 6 |
| | >=115 | 14 | 11 |
| Number of inpatient visits in the previous year | <1 | 0 | 0 |
| | [1, 4) | 12 | 9 |
| | >=4 | 23 | 20 |
| Metastatic cancer | No | -- | 0 |
| | Yes | -- | 19 |

"[A, B)" indicates an interval inclusive of the lower limit and exclusive of the upper limit. "--" indicates variables not included in a model.
h: hours; min: minutes; ED: Emergency department; LOS: length of hospital stay.

**Table 3.** Evaluation of prediction models on the test set, after fine-tuning cut-off values for continuous variables. The 95% CIs were generated from 100 bootstrap samples of the test set.

|  | Number of variables | mAUC (95% CI) | Generalized c-index (95% CI) |
|---|---|---|---|
| AutoScore-Ordinal Model 1* | 8 | 0.758 (0.754, 0.761) | 0.737 (0.734, 0.741) |
| POM1* | 8 | 0.750 (0.747, 0.754) | 0.726 (0.722, 0.729) |
| RF1* | 8 | 0.767 (0.764, 0.771) | 0.547 (0.544, 0.549) |
| AutoScore-Ordinal Model 2** | 8 | 0.793 (0.789, 0.796) | 0.760 (0.757, 0.763) |
| POM2** | 8 | 0.790 (0.786, 0.793) | 0.754 (0.750, 0.756) |
| RF2** | 8 | 0.798 (0.794, 0.801) | 0.564 (0.561, 0.566) |
| POM (stepwise) | 35 | 0.815 (0.812-0.819) | 0.775 (0.772-0.778) |
| POM (LASSO) | 10 | 0.704 (0.700-0.708) | 0.669 (0.665-0.673) |

*These models used the same 8 variables: ED LOS, creatinine, ED boarding time, number of visits in the previous year, age, systolic blood pressure (SBP), bicarbonate and pulse.
**These models used the same 8 variables: ED LOS, creatinine, number of visits in the previous year, age, systolic blood pressure (SBP), bicarbonate, pulse and metastatic cancer.
POM: proportional odds model; RF: random forest; mAUC: mean area under the curve.

**Interpreting prediction scores**

The AutoScore-generated score (from Models 1 and 2) can be mapped to the likelihood of falling into different outcome categories based on the observed proportions in the training set. For example, we illustrate the use of Model 2 for risk prediction for a hypothetical new patient in Figure 3. With values of the 8 variables measured for this new patient, clinicians can simply check relevant rows in the scoring table, summate the partial scores to a total score for this patient, and read the corresponding predicted probabilities for the three outcome categories in the lookup table. Such predicted probabilities can also be calculated from POM using a calculator or be returned from RF using designated software commands, but the checklist-style scoring table of AutoScore-generated models and the accompanying lookup tables of predicted probabilities are much easier to use in clinical practice.

**Figure 3.** Scoring and lookup tables for AutoScore-generated Model 2, with their use illustrated for a hypothetical new patient.

**Scoring table.**

| Variable | Value | Partial score | New patient |
|---|---|---|---|
| ED length of stay | <40min | 8 | |
| | [40min, 80min) | 6 | ✓ |
| | [80min, 4h) | 3 | |
| | [4h, 6h) | 1 | |
| | >=6h | 0 | |
| Creatinine, µmol/L | <45 | 4 | |
| | [45, 60) | 0 | ✓ |
| | [60, 135) | 1 | |
| | >=135 | 7 | |
| No. inpatient visits in the previous year | <1 | 0 | |
| | [1, 4) | 9 | |
| | >=4 | 20 | ✓ |
| Age (years) | <25 | 0 | |
| | [25, 45) | 4 | |
| | [45, 75) | 13 | ✓ |
| | [75, 85) | 17 | |
| | >=85 | 21 | |
| Systolic blood pressure, mmHg | <100 | 8 | |
| | [100, 120) | 4 | ✓ |
| | [120, 150) | 1 | |
| | >=150 | 0 | |
| Bicarbonate, mmol/L | <17 | 7 | |
| | [17, 20) | 3 | ✓ |
| | [20, 28) | 0 | |
| | >=28 | 4 | |
| Pulse, beat/minute | <70 | 0 | |
| | [70, 95) | 2 | ✓ |
| | [95, 115) | 6 | |
| | >=115 | 11 | |
| Metastatic cancer | No | 0 | ✓ |
| | Yes | 18 | |

**Total score = 6 + 0 + 20 + 13 + 4 + 3 + 2 + 0 = 48**

**Lookup table for predicted probabilities in the training set.**

| Total score | Outcome category 1 (alive, no 30-day readmission) | Outcome category 2 (alive, with 30-day readmission) | Outcome category 3 (death) |
|---|---|---|---|
| [0,5] | 0.941 | 0.059 | 0.000 |
| (5,10] | 0.965 | 0.033 | 0.002 |
| (10,15] | 0.962 | 0.035 | 0.003 |
| (15,20] | 0.940 | 0.053 | 0.007 |
| (20,25] | 0.900 | 0.076 | 0.023 |
| (25,30] | 0.855 | 0.107 | 0.038 |
| (30,35] | 0.793 | 0.147 | 0.060 |
| (35,40] | 0.718 | 0.190 | 0.092 |
| (40,45] | 0.626 | 0.240 | 0.134 |
| **(45,50]** | **0.545** | **0.289** | **0.166** |
| (50,55] | 0.472 | 0.246 | 0.281 |
| (55,60] | 0.427 | 0.227 | 0.346 |
| (60,65] | 0.306 | 0.216 | 0.478 |
| (65,70] | 0.254 | 0.199 | 0.547 |
| (70,100] | 0.186 | 0.212 | 0.602 |

Interval "[A, B)" is inclusive of the lower limit and exclusive of the upper limit.

We validated the lookup tables generated by Models 1 and 2 on the test set, visually presented in Figure 4, where the bars represent the observed proportions in the test set and the dots represent the predicted proportions based on the training set. An increase in the scores generally reflects an increased likelihood of being in a higher category in the outcome, whereas Model 2 has improved ability compared to Model 1 in differentiating different outcome categories given different predicted scores.

**Comparison with other approaches**

AutoScore-generated prediction models had comparable mAUC as the POM that used the same variables (see Table 3, where POM1 and POM2 correspond to Models 1 and 2 respectively). The RF using the same variables as Model 1 (see RF1 in Table 3) had a higher mAUC than Model 1, but when compared with Model 2 the advantage of the corresponding RF (see RF2 in Table 3) in terms of mAUC is less pronounced. AutoScore-generated models had slightly higher generalized c-index than the corresponding POMs, and both were higher than the corresponding RFs. In particular, the generalized c-index of RFs were much lower than the corresponding AutoScore-generated models or POMs, due to the use of predicted labels instead of numeric scores when evaluating the performance of RF.

When using traditional model building methods to build sparse POM, stepwise algorithm using AIC failed to work when starting from the null model (i.e., without any variable), and ended up selecting 35 variables when starting from the full model (i.e., including all 41 variables). Although this POM with 35 models had a high mAUC and generalized c-index (see POM (stepwise) in Table 3), it is difficult to use in practical settings. The LASSO approach selected 10 variables (i.e., ED LOS, gender, ED triage code, total number of ICU stays in the past year, admission type, $SpO_2$, SBP, bicarbonate, sodium and diabetes with complications) that had much lower performance than other models (see POM (LASSO) in Table 3).

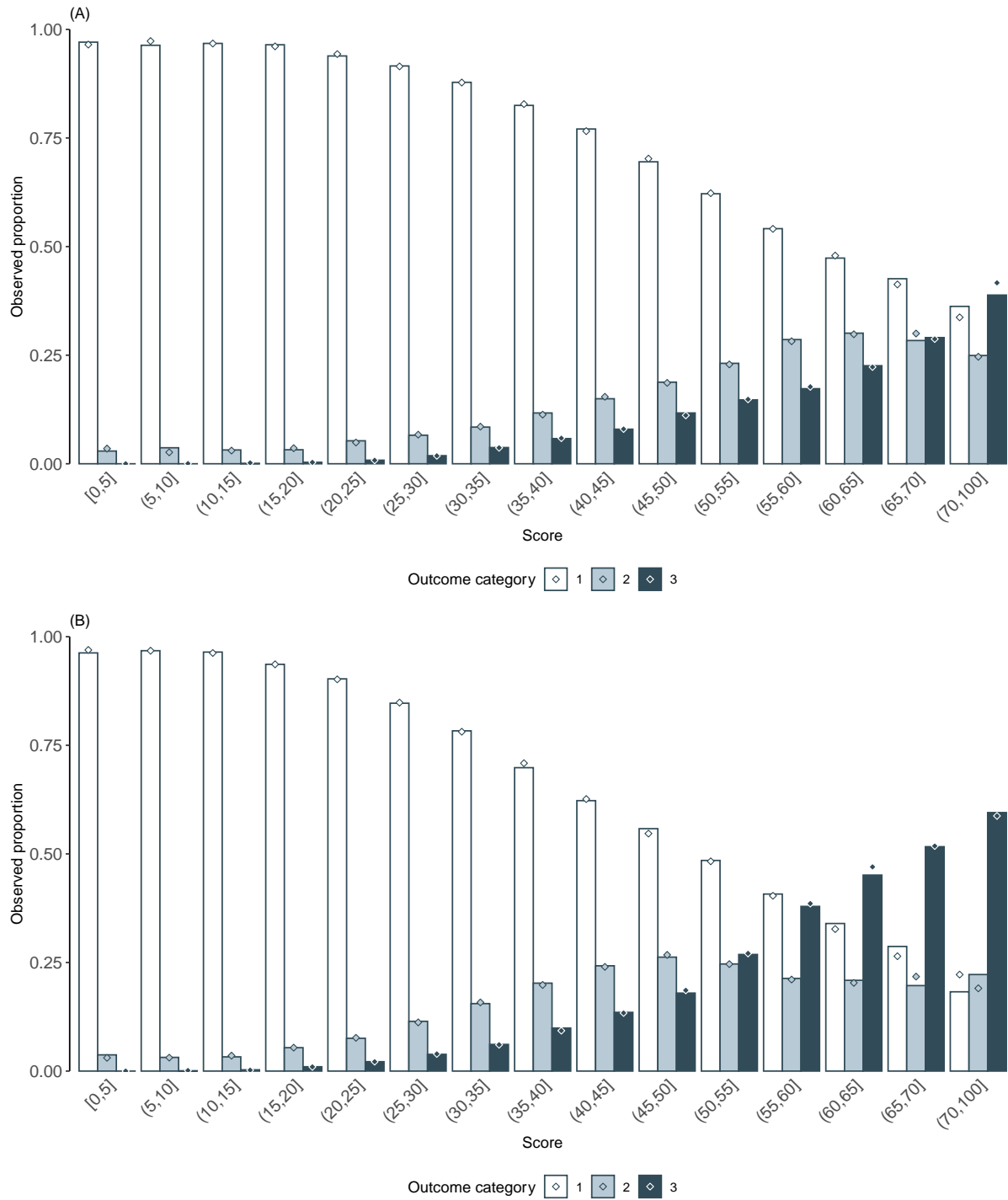

**Figure 4.** Proportion of cases observed in each outcome categories in the test set (bars) for cases with different AutoScore-generated prediction scores in (A) Model 1 and (B) Model 2. Points indicate the predicted proportions based on the training set.

**DISCUSSION**

A scoring system was developed using the AutoScore framework for ordinal outcomes in this study. The algorithm was applied on a case study to discuss the risk prediction model and its application on EHR data from the emergency department where the ordinal outcome includes three categories (alive without readmission to the hospital within 30 days post discharge, alive with readmission within 30 days post discharge and dead inpatient or within 30 days post discharge). The model was developed using 70% of the data (n=312,193); validated on subset of 10% of the data (n=44,599) to perform necessary model fine-tuning; and tested on a set of 20% (n=89,197). The performance of the AutoScore-Ordinal model was checked against the alternative models including POM and RF using 100 bootstrap samples via mAUC and generalized c-index. The AutoScore-Ordinal identified two feasible scoring models with 8 variables, and both had slightly better performance than the POM and RF that use the same variables. The novelty of the AutoScore-Ordinal model is its easy-to-use and machine learning-based automatic clinical score generator features, which develops interpretable clinical scoring models and can be useful tools for clinical decision-making at different stages of clinical pathway.

Prediction models in clinical settings are useful tools to inform clinical decision-making at different stages of clinical practice (44,45). To design, conduct and build prediction models, fundamental concepts including developing, validating and updating risk prediction models are discussed in the TRIPOD (Transparent Reporting of a multivariable prediction model for Individual Prognosis Or Diagnosis) Statement (46). New risk models should always be validated to quantify the predictive ability of the model (for example, calibration and discrimination), which could be addressed via internal (bootstrapping, cross-validation, etc.) or external (independent cohort, for example) validation (46).

Statistical methods and machine learning techniques have been used for ordinal outcomes in the literature. Multinomial logistic regression, random forest, eXtreme gradient boosting and gradient-boosted decision tree in a cross-sectional study on ED patients (47), ordinal classification method in a retrospective cohort study of ovarian cancer patients (27), multiple layer perceptron and random forest in a study across 9 mental health and suicide-related sub-Reddits (48), and multivariable POM in middle ear dysfunction diagnosis of infants (49) and in a coronary artery disease study have been explored in specific clinical domains (50). However there is a lack of interpretability and accessibility using these machine learning approaches. In contrast, the AutoScore-Ordinal via a point-based risk prediction model can be easily implemented in different clinical settings and fills a gap in interpretability, when dealing with ordinal outcomes.

The advantages of the original AutoScore framework (15) applies to the AutoScore-Ordinal framework. AutoScore-Ordinal builds on the POM, which is suitable for analyzing ordinal outcomes and widely used in clinical and epidemiological research. Compared to conventional use of POM, AutoScore-Ordinal makes use of machine learning methods to build sparse prediction models with good prediction performance, whereas traditional approaches such as stepwise variable selection and LASSO may not work well. AutoScore-Ordinal creates a check-list style scoring model that is easily implemented in clinical settings. In clinical research, sometimes quantitative data are categorized as ordinal variables due to different reasons such as skewness or multi-modal distribution. Under such scenarios, dichotomization may not be ideal and could result in loss of clinically and statistically relevant information. One may take advantage of the AutoScore-Ordinal framework to deal with such ordinal outcome variables.

AutoScore-Ordinal provides an efficient, straightforward and flexible variable selection procedure based on the parsimony plot, which visually presents the improvement in model performance with a growing number of variables in the model. Intuitively, researchers can select the top few variables that correspond to a satisfying model performance and inclusion of an additional variable results in a small (e.g., <1%) improvement, which resulted in Model 1 in our example. In addition, AutoScore-Ordinal allows researchers to manually add or remove variables from the final variables based on their contribution to model performance (e.g., as illustrated in Model 2) or practical implications.

POM is a member of a wider class of models, i.e., cumulative link models (30). While logit link is widely used in clinical applications, other link functions (e.g., probit, complementary log-log) may also be considered. AutoScore-Ordinal can be used with various link functions with minor modifications, and researchers may want to draw multiple parsimony plots to select a link function that best suits the data when determining variables to include in the final model.

In our data example we trained RF with 100 trees when ranking variables in Module 1 of AutoScore-Ordinal and when using it as a prediction model. Researchers may want to increase the number of trees to improve performance in general applications, e.g., 500 trees is a common choice (51). Due to the large sample size of our case study, we run out of memory when training an RF with 500 trees, and an RF with 200 trees generated comparable results when ranking variables and predicting ordinal outcomes.

As indicated by the name, POM assumes proportional odds, i.e., the effect of each variable on the outcome is the same across outcome categories. In univariable POM analyses of the training set (without categorizing continuous variables), the proportional odds assumption was rejected for all variables (with significance level of 5%). Future study should investigate how to relax this assumption when necessary without considerably complicating the interpretation of the resulting scoring model. Despite this, the two prediction models built using AutoScore-Ordinal worked reasonably well. For performance evaluation, we considered two metrics (i.e., mAUC and generalized c-index) that have straightforward interpretation and similar definition with metrics for binary and survival predictions (31,32,34). Future work may consider other performance metrics, e.g., volume under the receiver operating characteristic surface (more generally the hypervolume under the manifold) (52) and the ordinal c-index (31) for ordinal prediction, or the M-index (53) and polytomous discrimination index (54,55) for multi-class outcomes without explicitly accounting for ordering of categories.

Our data example aims to illustrate the use of our proposed AutoScore-Ordinal framework. The prediction performance can be improved, e.g., although Model 2 had better performance than Model 1, it will most likely fail to predict any new case into category 2, as this category is dominated by the other two categories (see Figure 4(B)). The AutoScore-Ordinal should be applied in other clinical domains with different sample sizes and various number of variables to establish external validity. Further investigation is required to improve performance before applying the AutoScore-Ordinal-derived scoring models in clinical settings, e.g., inclusion of additional relevant variables and alternative imputation of missing values. Nonetheless, AutoScore-Ordinal provides a powerful, flexible and easy-to-use framework for developing interpretable scoring models for ordinal clinical outcomes.

**CONCLUSION**

AutoScore-Ordinal as a risk prediction model was developed for ordinal outcome variable. For illustration purpose, the framework was implemented and validated using EHR data from the emergency department, where the ordinal outcome included three categories (alive without readmission to the hospital within 30 days post discharge, alive with readmission within 30 days post discharge and dead inpatient or within 30 days post discharge). An efficient and flexible variable selection procedure was explained and the model indicated a comparable goodness-of-fit in compared to the alternative models. The point-based risk prediction model generated by the AutoScore-Ordinal is easy to implement and interpret in different clinical settings.